\documentclass{book}
\usepackage{roboto}

\usepackage[english]{babel}
\usepackage[letterpaper,top=2cm,bottom=2cm,left=3cm,right=3cm,marginparwidth=1.75cm]{geometry}

\usepackage{pdflscape}
\usepackage{lscape}

\usepackage{rotating}
\usepackage{longtable}
\usepackage{array}
\usepackage{float}
\usepackage{amsmath}
\usepackage{graphicx}
\usepackage{inconsolata}
\usepackage[colorlinks=true, allcolors=blue]{hyperref}
\usepackage{enumitem}

\usepackage{tikz} 

\usepackage{listings}
\usepackage{xcolor}
\usepackage{epigraph}  

\usepackage[T1]{fontenc}
\usepackage{inconsolata}  

\definecolor{bgcolor}{rgb}{0.97,0.97,0.97}
\definecolor{codeblue}{rgb}{0.1,0.1,0.8}
\definecolor{codegreen}{rgb}{0,0.4,0}
\definecolor{codegray}{rgb}{0.4,0.4,0.4}
\definecolor{codepurple}{rgb}{0.5,0,0.5}
\definecolor{codered}{rgb}{0.6,0.2,0.2}
\definecolor{lightgray}{rgb}{0.9,0.9,0.9}
\definecolor{darkgray}{rgb}{0.6,0.6,0.6} 

\makeatletter
\renewcommand{\paragraph}{%
  \@startsection{paragraph}{4}{\z@}{1ex}{-1em}{\normalfont\normalsize\bfseries\color{gray}}}
\makeatother

\lstdefinestyle{python}{
    language=Python,
    basicstyle=\ttfamily\small\color{black}\usefont{T1}{zi4}{m}{n},  
    keywordstyle=\bfseries\color{codeblue},  
    stringstyle=\color{codegreen},  
    commentstyle=\slshape\color{codegray},  
    showstringspaces=false,
    numbers=left,
    numberstyle=\tiny\color{codegray},  
    stepnumber=1,
    numbersep=8pt,
    frame=single,
    rulecolor=\color{darkgray},  
    breaklines=true,
    backgroundcolor=\color{bgcolor},
    tabsize=4,
    captionpos=b,
    morekeywords={self}, 
}

\lstdefinestyle{text}{
    language=,
    basicstyle=\ttfamily\small\color{black}\usefont{T1}{zi4}{m}{n},  
    stringstyle=\color{codered},
    commentstyle=\color{codegray},
    showstringspaces=false,
    numbers=none,
    frame=single,
    rulecolor=\color{lightgray},  
    frameround=tttt,
    breaklines=true,
    backgroundcolor=\color{bgcolor},
    tabsize=4,
    captionpos=b,
}

\lstdefinestyle{cmd}{
    language=bash,
    basicstyle=\ttfamily\small\color{black}\usefont{T1}{zi4}{m}{n},  
    keywordstyle=\bfseries\color{blue},
    stringstyle=\color{codegreen},
    commentstyle=\itshape\color{gray},
    showstringspaces=false,
    numbers=none,
    frame=single,
    rulecolor=\color{darkgray},  
    breaklines=true,
    backgroundcolor=\color{bgcolor},
    tabsize=4,
    captionpos=b,
}

\lstdefinestyle{sql}{
    language=SQL,  
    basicstyle=\ttfamily\small\color{black}\usefont{T1}{zi4}{m}{n},  
    keywordstyle=\bfseries\color{codeblue},  
    stringstyle=\color{codegreen},  
    commentstyle=\slshape\color{codegray},  
    showstringspaces=false,  
    numbers=left,  
    numberstyle=\tiny\color{codegray},  
    stepnumber=1,  
    numbersep=8pt,  
    frame=single,  
    rulecolor=\color{darkgray},  
    breaklines=true,  
    backgroundcolor=\color{bgcolor},  
    tabsize=4,  
    captionpos=b,  
    morekeywords={
        SELECT, INSERT, DELETE, UPDATE, FROM, WHERE, AND, OR, JOIN, ON, CREATE, INDEX, TABLE, VALUES, INTO, AS, DISTINCT, ORDER, BY, GROUP, HAVING, LIMIT, OFFSET, UNION, DROP, ALTER, TRUNCATE, RENAME, PRIMARY, FOREIGN, KEY, CONSTRAINT, NULL, NOT, DEFAULT, AUTO_INCREMENT, UNIQUE, CHECK,
        db, find, insertOne, updateOne, deleteOne, collection, aggregate, match, project, sort, limit, pipeline, insertMany, updateMany, deleteMany, findOne, $lookup, $group,
        GET, SET, DEL, HGET, HSET, HDEL, LPUSH, RPUSH, LPOP, RPOP, SADD, SREM, PUBLISH, SUBSCRIBE, EXPIRE, TTL, FLUSHDB, FLUSHALL, INCR, DECR,
        SELECT, INSERT, UPDATE, DELETE, FROM, WHERE, AND, OR, USE, KEYSPACE, CREATE, ALTER, DROP, TRUNCATE, TABLE, INDEX, PRIMARY, FOREIGN, KEY, WITH, CLUSTERING, ORDER, BY, ASC, DESC, LIMIT, BATCH, APPLY, TOKEN, CONSISTENCY, QUORUM, LOCAL_QUORUM, ANY, ALL, ONE, TWO, THREE,
        MATCH, CREATE, MERGE, RETURN, DELETE, DETACH, REMOVE, SET, FOREACH, UNWIND, WITH, ORDER, BY, ASCENDING, DESCENDING, SKIP, LIMIT, UNION, ALL, OPTIONAL, DISTINCT, WHERE, AND, OR, IN, STARTS, ENDS, CONTAINS, EXISTS, IS, NULL,
        SELECT, INSERT, DELETE, UPDATE, FROM, WHERE, AND, OR, JOIN, ON, CREATE, INDEX, TABLE, VALUES, INTO, AS, DISTINCT, ORDER, BY, GROUP, HAVING, LIMIT, OFFSET, UNION, DROP, ALTER, TRUNCATE, RENAME, PRIMARY, FOREIGN, KEY, CONSTRAINT, NULL, NOT, DEFAULT, AUTO_INCREMENT, UNIQUE, CHECK, SEQUENCE, SYNONYM, PACKAGE, FUNCTION, PROCEDURE, TRIGGER, VIEW, GRANT, REVOKE, COMMIT, ROLLBACK,
        SELECT, INSERT, DELETE, UPDATE, FROM, WHERE, AND, OR, JOIN, ON, CREATE, INDEX, TABLE, VALUES, INTO, AS, DISTINCT, ORDER, BY, GROUP, HAVING, LIMIT, OFFSET, UNION, DROP, ALTER, TRUNCATE, RENAME, PRIMARY, FOREIGN, KEY, CONSTRAINT, NULL, NOT, DEFAULT, AUTO_INCREMENT, UNIQUE, CHECK, SERIAL, BIGSERIAL, RETURNING, DO, LANGUAGE, PLPGSQL, BEGIN, END, IMMUTABLE, VOLATILE,
        SELECT, INSERT, DELETE, UPDATE, FROM, WHERE, AND, OR, JOIN, ON, CREATE, INDEX, TABLE, VALUES, INTO, AS, DISTINCT, ORDER, BY, GROUP, HAVING, LIMIT, OFFSET, UNION, DROP, ALTER, TRUNCATE, RENAME, PRIMARY, FOREIGN, KEY, CONSTRAINT, NULL, NOT, DEFAULT, AUTO_INCREMENT, UNIQUE, CHECK, INCREMENT, ENGINE, CHARSET, COLLATE, COMMENT
    }  
}

\lstdefinestyle{html}{
    language=HTML,
    basicstyle=\ttfamily\small\color{black}\usefont{T1}{zi4}{m}{n},  
    keywordstyle=\bfseries\color{codeblue},  
    stringstyle=\color{codegreen},  
    commentstyle=\slshape\color{codegray},  
    showstringspaces=false,
    numbers=left,
    numberstyle=\tiny\color{codegray},  
    stepnumber=1,
    numbersep=8pt,
    frame=single,
    rulecolor=\color{darkgray},  
    breaklines=true,
    backgroundcolor=\color{bgcolor},
    tabsize=4,
    captionpos=b,
    morekeywords={<!DOCTYPE,html,head,body,div,span,a,img,href,src,script,style}, 
}

\title{Deep Learning and Machine Learning: Advancing Big Data Analytics and Management with Object-Oriented Programming
}
\author{
    Tianyang Wang\textsuperscript{*} \\ 
    \textit{Xi’an Jiaotong-Liverpool University} \\
    Tianyang.Wang21@student.xjtlu.edu.cn
    \and
    Ziqian Bi \textsuperscript{*$\dagger$} \\
    \textit{Indiana University} \\
    bizi@iu.edu
    \and
    Sen Zhang \\ 
    \textit{Rutgers University} \\
    sen.z@rutgers.edu
    \and
    Jiawei Xu \\ 
    \textit{Purdue University} \\
    xu1644@purdue.edu
    \and
    Qian Niu \\ 
    \textit{Kyoto University} \\
    niu.qian.f44@kyoto-u.jp
    \and
    Benji Peng \\ 
    \textit{AppCubic} \\
    benji@appcubic.com
    \and
    Ming Li \\ 
    \textit{Georgia Institute of Technology} \\
    mli694@gatech.edu
    \and
    Yizhu Wen \\
    \textit{University of Hawaii} \\
    yizhuw@hawaii.edu
    \and
    Hsieh Weiche \\
    \textit{National Tsing Hua University} \\
    s112033645@m112.nthu.edu.tw 
    \and
    Junyu Liu \\ 
    \textit{Kyoto University} \\
    liu.junyu.82w@st.kyoto-u.ac.jp
    \and
    Yichao Zhang \\
    \textit{The University of Texas at Dallas} \\
    yichao.zhang.us@gmail.com
    \and
    Keyu Chen\\ 
    \textit{Georgia Institute of Technology} \\
    kchen637@gatech.edu
    \and
    Jinlang Wang \\ 
    \textit{University of Wisconsin-Madison} \\
    jinlang.wang@wisc.edu
    \and
    Xuanhe Pan \\ 
    \textit{University of Wisconsin-Madison} \\
    xpan73@wisc.edu
    \and
    Pohsun Feng \\
    \textit{National Taiwan Normal University} \\
    41075018h@ntnu.edu.tw
    \and
    Xinyuan Song \\
    \textit{Emory University} \\
    songxinyuan@pku.edu.cn
    \and
    Chia Xin Liang \\
    \textit{JTB Technology Corp.} \\
    cxldun@gmail.com
    \and
    Xinyuan Song \\
    \textit{Emory Unversity} \\
    xsong30@emory.edu
    \and
    Ming Liu \textsuperscript{$\dagger$} \\ 
    \textit{Purdue University} \\
    liu3183@purdue.edu
}
\date{}

\begin{document}

\maketitle

\begingroup
\renewcommand\thefootnote{}\footnote{
    \textsuperscript{*} Equal contribution \\
    \textsuperscript{$\dagger$} Corresponding author
}
\addtocounter{footnote}{0}
\endgroup

\epigraph{"The best way to predict the future is to invent it."}{\textit{Arthur C. Clarke}}

\epigraph{"Programming is the art of telling another human being what one wants the computer to do."}{\textit{Donald Knuth}}

\epigraph{"Programming is the art of algorithm design and the craft of debugging errant code."}{\textit{Ellen Ullman}}

\epigraph{"No matter which field of work you want to go in, it is of great importance to learn at least one programming language."}{\textit{Ram Ray}}

\tableofcontents  

\setcounter{part}{2}
\part{Advancing Your Skills}

\chapter{Introduction to Advancing Your Skills}

Congratulations! After completing the introductory tutorials, you now have a solid understanding of the basics of deep learning. You've learned about fundamental neural network structures, training methods, and have gained some confidence in applying these techniques to simple tasks. However, the world of deep learning is vast \cite{niu2024large}, and what you’ve seen so far is just the beginning. There are more powerful models and algorithms that will help you tackle the complex problems we encounter in real-world applications.

In this section, we will guide you deeper into the exciting world of deep learning and help you further advance your skills. We will begin with a deep dive into advanced classifiers, discussing key architectures such as ResNet \cite{he2016identity}, which have become fundamental in fields like computer vision. You’ll learn how these models work, focusing on the key innovations, such as residual networks, that address issues like vanishing gradients and overfitting in deep networks.

Next, we will explore the world of object detection. Object detection goes beyond simply identifying whether an object exists in an image; it also involves locating where these objects are. This task is critical in fields like autonomous driving and smart surveillance. We'll explore powerful models such as YOLO \cite{jiang2022review} and Faster R-CNN \cite{jiang2017face}, equipping you with the tools needed to tackle complex visual tasks that require object localization \cite{Peng_2024}.

As models become more sophisticated, your understanding of the underlying mathematics becomes increasingly important. Therefore, we will dedicate some time to strengthening and expanding your mathematical foundation. This will include advanced topics in calculus and linear algebra, as well as optimization techniques like gradient descent and learning rate adjustment. By mastering these concepts, you'll gain a deeper understanding of what drives model training and improve your ability to fine-tune your models.

In addition, managing and organizing your deep learning projects is an essential skill as you grow in this field. We’ll discuss the importance of project management, helping you efficiently handle code, data, models, and experiment results. Having strong project management habits allows you to stay focused on innovation and model optimization, especially in larger-scale projects.

Finally, we’ll wrap up with a discussion on object-oriented programming (OOP) \cite{rentsch1982object} in deep learning. As your projects grow in complexity, writing clean, modular, and reusable code becomes essential. We'll explore how OOP principles can help you improve code readability, maintainability, and scalability, making your development process more efficient.

In summary, this section is another key milestone in your deep learning journey. By mastering these advanced techniques and concepts, you’ll be able to build more powerful models and confidently tackle complex real-world problems. Ready to dive in? Let’s continue our deep learning adventure and reach the next level!

\chapter{Object-Oriented Programming (OOP)}

\section{The History of Computing}

\subsection{The Origins of Computing}
The origins of computing can be traced back to the early 19th century, long before the existence of electronic computers. The British mathematician Charles Babbage designed the first mechanical computer, the “Difference Engine \cite{ansell1997deleuze},” followed by the more ambitious “Analytical Engine \cite{bromley1982charles}.” Although these machines were never fully built during Babbage's time, their conceptual frameworks laid the groundwork for modern computers. 

The Analytical Engine was notable because it was programmable, thanks to input using punched cards \cite{harlizius2017weaving}—a technique borrowed from the Jacquard loom. Ada Lovelace, Babbage’s collaborator, is widely recognized as the world’s first programmer. She wrote what is considered the first algorithm intended for a machine, understanding that computers could be used for more than just calculations. This visionary perspective paved the way for modern computing.

\subsection{The Rise of Electronic Computers: World War II and Beyond}
The development of computing accelerated during World War II. The demand for rapid and reliable calculations, especially for ballistics and code-breaking, led to the construction of early electronic computers. Machines like the British “Colossus \cite{dell2014colossus}” and the American “ENIAC \cite{weik1961eniac}” were built during this period. ENIAC was used to calculate artillery firing tables for the U.S. Army and solve complex mathematical problems.

One notable aspect of this era was the significant role women played in programming. Many of the early programmers, including those who worked on ENIAC, were women, such as Jean Jennings and Frances Bilas \cite{todd2015jean,fritz1996women}. They wrote programs using punched cards and physical switches, which were highly labor-intensive but a vital step toward modern software development.

\subsection{Punched Cards and Early Programming}
Punched cards \cite{casey1951punched} were an essential input method in early computing. Each punched card represented a line of code or a specific instruction. These cards were fed into the computer in the correct sequence, allowing the machine to execute complex tasks. While punched cards made it possible to store and process information, they also had limitations. Debugging errors involved physically examining and reordering cards, making the process slow and error-prone. Despite these challenges, this method of programming was foundational, and it shaped early computing practices.

\section{The Evolution of Programming Languages}

\subsection{The Birth of High-Level Languages}
In the 1950s, computers became more powerful, and programming languages began evolving to make the process more abstract and efficient. Instead of directly interacting with hardware or writing in assembly language, high-level programming languages allowed developers to express commands in a more human-readable form.

Two major programming languages emerged in this era:
\begin{itemize}
    \item \textbf{Fortran (1957) \cite{backus1978history}}: Designed by IBM for scientific and engineering applications, Fortran (Formula Translation) allowed complex mathematical formulas to be written and computed easily. It was a major step forward in making programming more accessible to scientists and engineers.
    \item \textbf{COBOL (1959) \cite{sammet1978early}}: COBOL (Common Business-Oriented Language) was created for business and administrative tasks. Its syntax was designed to be close to English, enabling business professionals to understand code without specialized training.
\end{itemize}

These languages significantly improved the productivity of programmers and marked the beginning of an era in which software could be developed more quickly and on a larger scale.

\subsection{Procedural Programming}
By the 1960s and 1970s, the programming paradigm shifted toward procedural programming \cite{wegner1990concepts}. This approach involved writing software as a series of functions or procedures that performed specific tasks. Each function had a clear input-output mechanism, making the software more modular and easier to maintain. C, developed in 1972, became the most influential procedural programming language, as it allowed precise control over hardware while maintaining higher-level abstractions for logic and flow control.

However, as software systems grew in complexity, procedural programming began to encounter limitations. Functions often became too interdependent, and managing shared data between functions became cumbersome. These challenges sparked the development of more sophisticated programming paradigms.

\subsection{The Development of Computer Hardware}
During this time, the rapid advancement in computer hardware played a crucial role in shaping programming techniques. The invention of transistors and integrated circuits in the 1950s and 1960s allowed computers to become smaller, faster, and more reliable. By the 1970s, personal computers like the Apple II and the IBM PC emerged, bringing computing power to a much larger audience \cite{hagedoorn2001strange}. These advances enabled more sophisticated software development, as more memory and processing power became available for complex programs.

\section{The Rise of Object-Oriented Programming}

\subsection{The Basic Concepts of Object-Oriented Programming}
As computer systems grew more complex, a new programming paradigm called Object-Oriented Programming (OOP) emerged in the late 1970s and early 1980s \cite{nelson1990introduction,wegner1990concepts}. OOP’s core idea was to structure programs around “objects,” which encapsulate both data (attributes) and behavior (methods). Objects are instances of “classes,” which define a blueprint for creating similar objects \cite{pokkunuri1989object}.

The four core principles of OOP are:
\begin{itemize}
    \item \textbf{Encapsulation}: This principle involves bundling data and methods that operate on the data into a single unit, or object. Encapsulation helps hide the internal state of an object and only exposes necessary functionalities to the outside world.
    \item \textbf{Inheritance}: Inheritance allows new classes to inherit properties and behaviors from existing classes. This promotes code reuse and enables hierarchical relationships between classes.
    \item \textbf{Polymorphism}: Polymorphism allows objects to be treated as instances of their parent class. This enables different objects to respond to the same function call in different ways, depending on their specific implementation.
    \item \textbf{Abstraction}: Abstraction simplifies complex systems by modeling only the relevant aspects. It hides unnecessary details and exposes only the essential features needed to interact with the object.
\end{itemize}

OOP was designed to better model real-world problems and manage large software projects by breaking them down into smaller, reusable components.

\subsection{Early Object-Oriented Languages}
The first programming language to support OOP was \textbf{Simula} (1967) \cite{nygaard1978development}. Simula introduced the concept of classes and objects to simulate real-world systems. This was followed by \textbf{Smalltalk} (1972) \cite{kay1996early}, which further refined the OOP model and is considered one of the first fully object-oriented programming languages. Smalltalk introduced the idea that "everything is an object," including primitive data types like numbers and strings.

\subsection{The Popularity of C++ and Java}
The popularity of OOP grew in the 1980s with the development of \textbf{C++}, which extended the C programming language with object-oriented features \cite{stroustrup2007evolving}. C++ became widely adopted due to its combination of procedural and object-oriented capabilities. 

In the 1990s, \textbf{Java} was introduced, designed to be fully object-oriented and platform-independent. Java's slogan, "Write once, run anywhere," reflected its ability to run on any device with a Java Virtual Machine (JVM) \cite{karakaya2008java}, further promoting the OOP model. Java has since become one of the most widely used programming languages for building enterprise applications.

\section{Modular Programming and Design Patterns}

\subsection{Modular Programming}
As software systems grew in size and complexity, the need for better organization became apparent. Modular programming, which emphasizes the division of software into independent, interchangeable modules, emerged as a solution. Each module in a program handles a specific piece of functionality and interacts with other modules through well-defined interfaces.

The advantages of modular programming include:
\begin{itemize}
    \item \textbf{Enhanced Maintainability}: Developers can work on different parts of the system independently, making it easier to identify and fix bugs or extend functionality.
    \item \textbf{Code Reusability}: Well-defined modules can be reused across different projects or systems, reducing development time.
    \item \textbf{Team Collaboration}: Modular programming enables parallel development, where different teams can work on different modules simultaneously.
\end{itemize}

\subsection{Design Patterns}
Design patterns \cite{coplien1998software} are proven solutions to common software design problems. They offer a standardized approach to solving problems that arise frequently during software development. The most commonly used design patterns include:
\begin{itemize}
    \item \textbf{Singleton Pattern}: Ensures that a class has only one instance and provides a global point of access to that instance.
    \item \textbf{Factory Pattern}: Provides an interface for creating objects in a superclass, but allows subclasses to alter the type of objects that will be created.
    \item \textbf{Observer Pattern}: Establishes a one-to-many relationship between objects, where changes to one object are automatically reflected in dependent objects.
\end{itemize}

These patterns help developers create more flexible, reusable, and maintainable code by providing tested solutions to recurring problems. 

\section{Python and Object-Oriented Programming}

\subsection{The Story of Python}
Python was created in 1989 by Guido van Rossum with the intention of making a programming language that was easy to read and write \cite{van2007python}. Python was designed to bridge the gap between scripting and system programming languages by offering both power and simplicity. Python's syntax emphasizes readability, which makes it an excellent choice for both beginners and experienced developers. Over the years, Python has grown into a versatile language used for web development, data science, machine learning, and more\cite{chen2024deeplearningmachinelearning}.

\subsection{Object-Oriented Programming in Python}
Python is a multi-paradigm language, meaning it supports multiple programming styles, including procedural, functional, and object-oriented programming. In Python, everything is an object, including primitive data types like integers and strings. Python allows developers to create their own classes and define attributes and methods for those classes\cite{peng2024securinglargelanguagemodels}.

Here is a simple example of object-oriented programming in Python:

\begin{lstlisting}[style=python]
class Animal:
    def __init__(self, name):
        self.name = name

    def speak(self):
        raise NotImplementedError("Subclass must implement abstract method")

class Dog(Animal):
    def speak(self):
        return f"{self.name} says Woof!"

class Cat(Animal):
    def speak(self):
        return f"{self.name} says Meow!"

dog = Dog("Buddy")
cat = Cat("Kitty")

print(dog.speak())  # Output: Buddy says Woof!
print(cat.speak())  # Output: Kitty says Meow!
\end{lstlisting}

In this example, the \texttt{Animal} class serves as an abstract base class, and \texttt{Dog} and \texttt{Cat} are subclasses that inherit from it. Each subclass implements the \texttt{speak} method to provide specific behavior.

\subsection{Modularity and Design Patterns in Python}
Python supports modular programming through its extensive module and package system. Developers can organize code into modules (files) and packages (directories), making it easy to reuse functionality across projects. Python’s standard library contains a wide range of modules that help developers handle tasks like file I/O, networking, and database access.

Python also supports the implementation of design patterns. For instance, the Singleton pattern can be easily implemented in Python using modules or classes. Similarly, Python’s dynamic typing and flexible class system make it well-suited for implementing design patterns like the Factory and Observer patterns.

\section{Object-Oriented Programming in Artificial Intelligence}
Object-Oriented Programming (OOP) is a powerful programming paradigm that structures code using objects and classes. This approach enables efficient and organized development, making complex systems like machine learning (ML)\cite{Peng_2024}, deep learning (DL), large language models (LLM)\cite{niu2024large}, and data analytics more manageable, reusable, and scalable\cite{peng2024deeplearningmachinelearning}.

In machine learning, encapsulation is useful for wrapping preprocessing steps, models, and evaluation metrics into objects. Here’s an example using Python’s \texttt{scikit-learn} library to encapsulate the process of training a linear regression model.

\begin{lstlisting}[style=python]
from sklearn.linear_model import LinearRegression
from sklearn.model_selection import train_test_split
from sklearn.metrics import mean_squared_error

# Define the Machine Learning Model class
class MLModel:
    def __init__(self, model):
        self.model = model

    # Encapsulation of training process
    def train(self, X, y):
        self.model.fit(X, y)

    # Encapsulation of prediction process
    def predict(self, X):
        return self.model.predict(X)

    # Encapsulation of evaluation process
    def evaluate(self, X_test, y_test):
        predictions = self.model.predict(X_test)
        mse = mean_squared_error(y_test, predictions)
        print(f"Mean Squared Error: {mse}")

# Example usage
if __name__ == "__main__":
    # Example data
    X = [[1], [2], [3], [4], [5]]
    y = [1, 2, 3, 4, 5]
    
    # Train-test split
    X_train, X_test, y_train, y_test = train_test_split(X, y, test_size=0.2)

    # Initialize the model and encapsulate logic
    model = MLModel(LinearRegression())
    model.train(X_train, y_train)
    model.evaluate(X_test, y_test)
\end{lstlisting}

In this example, the \texttt{MLModel} class encapsulates the linear regression model from \texttt{scikit-learn}. The methods \texttt{train}, \texttt{predict}, and \texttt{evaluate} handle training, predicting, and evaluating the model, keeping the code organized and reusable.

\section{Conclusion}
From the early days of mechanical computing to modern software development practices, the field of computing has evolved tremendously. Object-oriented programming has become a dominant paradigm due to its ability to model real-world entities and simplify complex software systems. As systems grow larger and more complex, modular programming and design patterns have emerged as essential tools for creating maintainable and scalable software architectures.

Python, as a versatile and easy-to-learn language, embraces OOP principles and modular design, making it an excellent choice for both small projects and large-scale applications. In the next chapters, we will explore practical examples of using Python’s object-oriented features and applying design patterns to create robust and efficient software systems.

\section{Introduction to OOP}
    \subsection{Introduction}
    Object-Oriented Programming (OOP) is a programming paradigm that structures a program by bundling related properties and behaviors into individual objects. Objects are the central concept of OOP. In Python, as in other object-oriented languages, OOP principles make it easier to model real-world entities by organizing code around these objects, each of which contains both data (attributes) and functionality (methods). 
    
    The four foundational pillars of OOP are:
    
    \begin{itemize}
        \item \textbf{Encapsulation}: Bundling data (attributes) and methods (functions) into a single unit (class). It also restricts direct access to certain attributes or methods from outside the class.
        \item \textbf{Inheritance}: Allows a new class (derived class) to inherit attributes and methods from an existing class (base class), promoting code reuse.
        \item \textbf{Polymorphism}: Allows objects of different types to be treated as objects of a common superclass, enabling different implementations of a method in different classes.
        \item \textbf{Abstraction}: Hides the complex implementation details and exposes only the necessary parts, making code simpler and more readable.
    \end{itemize}

    Let's explore these principles in greater detail through Python's syntax and practical examples.

\section{Classes and Objects}

    \subsection{Introduction}
    In Python, everything is an object. Every entity, be it a number, a string, or a data structure, is treated as an object in Python. To create a new type of object, we define a class. A class is a blueprint or template for creating objects, while an object is an instance of that class.

    To clarify the relationship between classes and objects:
    
    \begin{itemize}
        \item \textbf{Class}: A class is a user-defined data type that defines a set of attributes (variables) and methods (functions) that operate on that data. It acts as a template for creating objects.
        \item \textbf{Object}: An object is an instance of a class. When a class is defined, no memory is allocated until an object is instantiated from the class.
    \end{itemize}
    
    For example, think of a class as a blueprint for a car. The blueprint itself isn't a car, but you can use it to create multiple car objects.

    \subsection{Class Definition}
    In Python, a class is defined using the \texttt{class} keyword. Let’s look at how to define a simple class:
    
    \begin{lstlisting}[style=python]
class Car:
    # Class attribute
    wheels = 4

    # Constructor method (__init__): initializes the object
    def __init__(self, color, brand):
        # Instance attributes
        self.color = color
        self.brand = brand

    # Method to describe the car
    def describe(self):
        return f'This car is a {self.color} {self.brand} with {self.wheels} wheels.'
    \end{lstlisting}
    
    \begin{itemize}
        \item \texttt{class Car:} — This is how we define a class named \texttt{Car}.
        \item \texttt{wheels = 4} — This is a class attribute, shared by all instances of the class.
        \item \texttt{\_\_init\_\_} — This is a special method known as the constructor, called automatically when a new object is instantiated. It initializes the object's attributes.
        \item \texttt{self.color, self.brand} — These are instance attributes, unique to each object created from the class.
        \item \texttt{describe()} — This is a method (a function defined within a class) that describes the object using its attributes.
    \end{itemize}

    \subsection{Object Instantiation}
    Once a class is defined, we can create instances (objects) of that class. Instantiating an object means calling the class to create a new instance of it. Here's how we can instantiate objects from the \texttt{Car} class:
    
    \begin{lstlisting}[style=python]
# Creating objects from the Car class
car1 = Car("red", "Toyota")
car2 = Car("blue", "Honda")

# Calling the describe method on the objects
print(car1.describe())  # Output: This car is a red Toyota with 4 wheels.
print(car2.describe())  # Output: This car is a blue Honda with 4 wheels.
    \end{lstlisting}
    
    In this example:
    \begin{itemize}
        \item \texttt{car1 = Car("red", "Toyota")} creates a new \texttt{Car} object with color "red" and brand "Toyota".
        \item \texttt{car2 = Car("blue", "Honda")} creates another \texttt{Car} object with color "blue" and brand "Honda".
        \item \texttt{car1.describe()} and \texttt{car2.describe()} call the \texttt{describe()} method on each object, providing information about them.
    \end{itemize}

    \subsection{Attributes and Methods}
    Attributes and methods are what define the structure and behavior of an object:
    
    \begin{itemize}
        \item \textbf{Attributes}: These are the data associated with an object. There are two types of attributes:
            \begin{itemize}
                \item \textbf{Instance attributes}: Unique to each object. In the \texttt{Car} class, \texttt{color} and \texttt{brand} are instance attributes.
                \item \textbf{Class attributes}: Shared among all objects of the class. In the \texttt{Car} class, \texttt{wheels} is a class attribute, meaning all cars will have four wheels.
            \end{itemize}
        \item \textbf{Methods}: Functions defined within a class that operate on objects of that class. In our example, \texttt{describe()} is a method that returns a description of the car.
    \end{itemize}
    
    Here's an illustration to better understand the concept of classes and objects:

    
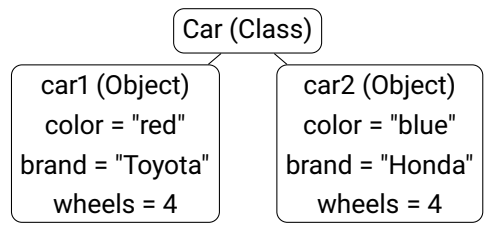
\begin{center}
    \begin{tikzpicture}[sibling distance=10em, every node/.style = {shape=rectangle, rounded corners, draw, align=center}]
      \node {Car (Class)}
        child { node {car1 (Object) \\ color = "red" \\ brand = "Toyota" \\ wheels = 4} }
        child { node {car2 (Object) \\ color = "blue" \\ brand = "Honda" \\ wheels = 4} };
    \end{tikzpicture}
    \end{center}

    As shown, both \texttt{car1} and \texttt{car2} are instances of the \texttt{Car} class, each having unique instance attributes (\texttt{color} and \texttt{brand}) but sharing the same class attribute (\texttt{wheels}).

\section{Encapsulation}

\subsection{Introduction}
Encapsulation \cite{skoglund2003practical} is a fundamental concept in Object-Oriented Programming (OOP) that involves bundling the data (attributes) and methods (functions) that operate on the data within a single unit or class. It is one of the core principles of OOP, along with inheritance, polymorphism, and abstraction.

In Python, encapsulation allows you to restrict direct access to some of an object's components, which is essential for protecting the data from accidental modification and misuse. Encapsulation is implemented through the use of classes. It allows you to define methods for setting and retrieving the values of an object's attributes, ensuring that they are not directly accessed or altered without proper control.

\subsection{Access Modifiers (Private, Public, Protected)}
Access modifiers \cite{jackson2016objects} are special keywords or symbols used to control the visibility of class attributes and methods. In Python, access modifiers are achieved using naming conventions. Python does not enforce strict access control as in some other languages like Java or C++, but it does provide mechanisms to indicate how a member of a class should be accessed.

\subsubsection{Public Members}
Public members are accessible from anywhere. In Python, attributes and methods are public by default. This means they can be accessed both from inside and outside the class.

\begin{lstlisting}[style=python]
class Dog:
    def __init__(self, name):
        self.name = name  # Public attribute

    def bark(self):
        print(f"{self.name} is barking!")

dog = Dog("Rex")
dog.bark()  # Rex is barking!
print(dog.name)  # Accessing public attribute directly
\end{lstlisting}

In the above example, the \texttt{name} attribute and the \texttt{bark} method are public, meaning they can be accessed and modified directly from outside the class.

\subsubsection{Private Members}
Private members are intended to be accessed only within the class where they are defined. In Python, we use a double underscore \texttt{\_\_} before the attribute or method name to indicate that it is private.

\begin{lstlisting}[style=python]
class Cat:
    def __init__(self, name):
        self.__name = name  # Private attribute

    def __meow(self):  # Private method
        print(f"{self.__name} is meowing!")

    def make_sound(self):
        self.__meow()  # Accessing private method within the class

cat = Cat("Whiskers")
cat.make_sound()  # Whiskers is meowing!
# print(cat.__name)  # AttributeError: 'Cat' object has no attribute '__name'
\end{lstlisting}

In this example, the attribute \texttt{\_\_name} and the method \texttt{\_\_meow} are private. They cannot be accessed directly from outside the class. Attempting to access them will result in an error. However, they can be accessed from within the class itself.

\subsubsection{Protected Members}
Protected members are intended to be accessed within the class and its subclasses. In Python, protected members are denoted by a single underscore \texttt{\_} before the attribute or method name. This indicates that the member is protected, but it is still accessible from outside the class, though it is a convention that such members should not be accessed directly.

\begin{lstlisting}[style=python]
class Animal:
    def __init__(self, species):
        self._species = species  # Protected attribute

    def _move(self):  # Protected method
        print(f"The {self._species} is moving.")

class Dog(Animal):
    def bark(self):
        print(f"The {self._species} is barking!")
        self._move()  # Accessing protected method from a subclass

dog = Dog("dog")
dog.bark()
\end{lstlisting}

Here, the attribute \texttt{\_species} and the method \texttt{\_move} are protected. They can be accessed from the subclass \texttt{Dog}, but should not generally be accessed from outside the class.

\subsection{Getters and Setters}
In Python, it is common practice to use \texttt{getter} and \texttt{setter} methods to control access to private attributes. These methods allow us to retrieve and update the values of private attributes in a controlled manner, ensuring that they are not accessed or modified in an unsafe way.

\begin{lstlisting}[style=python]
class Person:
    def __init__(self, name, age):
        self.__name = name  # Private attribute
        self.__age = age  # Private attribute

    # Getter for name
    def get_name(self):
        return self.__name

    # Setter for name
    def set_name(self, name):
        self.__name = name

    # Getter for age
    def get_age(self):
        return self.__age

    # Setter for age
    def set_age(self, age):
        if age > 0:
            self.__age = age
        else:
            print("Age cannot be negative!")

person = Person("Alice", 25)
print(person.get_name())  # Alice
person.set_age(30)
print(person.get_age())  # 30
person.set_age(-5)  # Age cannot be negative!
\end{lstlisting}

In this example, we use getter and setter methods for controlled access to the private attributes \texttt{\_\_name} and \texttt{\_\_age}. The setter for \texttt{age} includes a check to ensure that the age cannot be set to a negative value, which helps protect the integrity of the data.

\subsection{Benefits of Encapsulation}
Encapsulation offers several advantages, which make it a critical aspect of OOP:

\subsubsection{1. Data Protection}
Encapsulation protects the internal state of an object by preventing unauthorized or accidental access to sensitive data. By restricting access to the attributes and providing controlled methods to interact with them, encapsulation ensures that the data remains valid and consistent.

\subsubsection{2. Increased Security}
By hiding the implementation details of a class and exposing only the necessary methods, encapsulation ensures that an object’s internal workings are not visible to the outside world. This enhances security as it limits how external code can interact with the object's data.

\subsubsection{3. Maintainability}
Encapsulation improves maintainability by allowing you to change the internal implementation of a class without affecting the external code that uses it. If the implementation details change, you only need to update the class's methods without requiring changes to the code that interacts with the object.

\subsubsection{4. Flexibility and Modularity}
Encapsulation allows you to modify or extend the functionality of a class in the future without breaking existing code. This modular approach makes it easier to manage larger codebases by focusing on the specific functionalities of each class.

\subsubsection{5. Controlled Access}
With encapsulation, you have full control over how an object’s data is accessed and modified. By using getters and setters, you can ensure that only valid data is assigned to attributes, providing better control and validation.

Encapsulation plays a pivotal role in writing clean, modular, and maintainable code. By restricting access to data and providing a controlled interface for interaction, it enhances the robustness and flexibility of your Python programs.

\section{Inheritance}
\subsection{Introduction}
Inheritance \cite{cartwright1998empirical} is one of the fundamental concepts of Object-Oriented Programming (OOP) in Python. It allows one class to inherit attributes and methods from another class, promoting code reuse and establishing hierarchical relationships between classes. Inheritance helps to avoid redundancy and makes code more maintainable and scalable.

The class that is inherited from is called the "parent" or "base" class, and the class that inherits is referred to as the "child" or "derived" class. By using inheritance, a child class can access and use the attributes and methods of the parent class, as well as define its own unique properties.

\begin{lstlisting}[style=python]
# Example of inheritance
class Animal:
    def speak(self):
        return "Some generic sound"

class Dog(Animal):
    def speak(self):
        return "Bark"
        
dog = Dog()
print(dog.speak())  # Output: Bark
\end{lstlisting}

In the above example, the \texttt{Dog} class inherits from the \texttt{Animal} class. Although the \texttt{Dog} class defines its own \texttt{speak} method, it inherits all other methods and properties of the \texttt{Animal} class (if any).

\subsection{Single Inheritance}
Single inheritance occurs when a class inherits from one parent class only. This is the most straightforward form of inheritance. In Python, a class can directly inherit attributes and methods from a single base class, and it can also override methods from the parent class to provide specific implementations.

\begin{lstlisting}[style=python]
# Single Inheritance example
class Animal:
    def __init__(self, name):
        self.name = name
    
    def speak(self):
        return f"{self.name} makes a sound"

class Cat(Animal):
    def speak(self):
        return f"{self.name} says Meow"
        
cat = Cat("Whiskers")
print(cat.speak())  # Output: Whiskers says Meow
\end{lstlisting}

In this example, the \texttt{Cat} class is a child of the \texttt{Animal} class. The \texttt{Cat} class inherits the \texttt{\_\_init\_\_} method from the parent class, but it overrides the \texttt{speak} method to provide its own implementation.

\subsection{Multiple Inheritance}
Multiple inheritance occurs when a class inherits from more than one parent class. This allows a child class to inherit attributes and methods from multiple classes. While powerful, multiple inheritance can introduce complexity, particularly when there are methods with the same name in different parent classes.

\begin{lstlisting}[style=python]
# Multiple Inheritance example
class Animal:
    def move(self):
        return "Moves on land"

class Bird:
    def move(self):
        return "Flies in the air"

class Penguin(Animal, Bird):
    def move(self):
        return "Swims in water"
        
penguin = Penguin()
print(penguin.move())  # Output: Swims in water
\end{lstlisting}

In this example, the \texttt{Penguin} class inherits from both \texttt{Animal} and \texttt{Bird} classes, but it overrides the \texttt{move} method to provide its own unique behavior.

\paragraph{Method Resolution Order (MRO):} In multiple inheritance, Python uses the Method Resolution Order (MRO) to determine which parent class method should be used when a method is called. Python follows the C3 linearization algorithm to resolve the order of method calls. You can check the MRO of a class using the \texttt{\_\_mro\_\_} attribute or the \texttt{mro()} method.

\begin{lstlisting}[style=python]
# MRO example
print(Penguin.__mro__)
# Output: (<class '__main__.Penguin'>, <class '__main__.Animal'>, <class '__main__.Bird'>, <class 'object'>)
\end{lstlisting}

\subsection{Multilevel Inheritance}
Multilevel inheritance occurs when a class inherits from another derived class, forming a chain of inheritance. This creates a hierarchical structure, where each class can build upon the previous one.

\begin{lstlisting}[style=python]
# Multilevel Inheritance example
class Animal:
    def __init__(self, name):
        self.name = name
    
    def speak(self):
        return f"{self.name} makes a sound"

class Mammal(Animal):
    def __init__(self, name):
        super().__init__(name)

class Dog(Mammal):
    def speak(self):
        return f"{self.name} says Bark"

dog = Dog("Buddy")
print(dog.speak())  # Output: Buddy says Bark
\end{lstlisting}

Here, the \texttt{Dog} class inherits from the \texttt{Mammal} class, which in turn inherits from the \texttt{Animal} class. The \texttt{super()} function is used in the \texttt{Mammal} class to call the constructor of the \texttt{Animal} class, and the \texttt{Dog} class provides its own implementation of the \texttt{speak} method.

\subsection{Overriding Methods}
Method overriding allows a child class to provide a specific implementation for a method that is already defined in its parent class. This is particularly useful when a child class needs to modify or extend the behavior of the parent class method.

\begin{lstlisting}[style=python]
# Method Overriding example
class Animal:
    def speak(self):
        return "Animal makes a sound"

class Dog(Animal):
    def speak(self):
        return "Bark"

animal = Animal()
dog = Dog()
print(animal.speak())  # Output: Animal makes a sound
print(dog.speak())     # Output: Bark
\end{lstlisting}

In this case, the \texttt{Dog} class overrides the \texttt{speak} method inherited from the \texttt{Animal} class. Even though both classes have a \texttt{speak} method, the child class \texttt{Dog} uses its own version when called.

\subsection{Super Keyword}
The \texttt{super()} keyword is used in Python to refer to the parent class and is commonly used in method overriding to call methods from the parent class. This is especially useful when you want to extend the behavior of a parent method rather than completely replace it.

\begin{lstlisting}[style=python]
# Using super() in method overriding
class Animal:
    def speak(self):
        return "Animal makes a sound"

class Dog(Animal):
    def speak(self):
        parent_sound = super().speak()
        return f"{parent_sound} and Dog barks"

dog = Dog()
print(dog.speak())  # Output: Animal makes a sound and Dog barks
\end{lstlisting}

In this example, the \texttt{Dog} class uses \texttt{super()} to call the \texttt{speak} method of the \texttt{Animal} class, and then appends its own behavior to the result. This allows for extending functionality without losing the original method's behavior.

\subsection{Class Hierarchy Visualization}
To better understand the hierarchical relationships between classes in inheritance, let's visualize the class structure in a tree-like diagram using the \texttt{tikzpicture} package in \LaTeX.

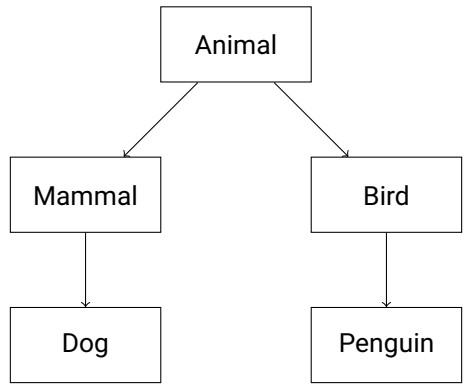
\begin{center}
\begin{tikzpicture}
    \tikzstyle{every node}=[draw, rectangle, minimum width=2cm, minimum height=1cm] 
    \node (A) at (0,0) {Animal};
    \node (B) at (-2,-2) {Mammal};
    \node (C) at (2,-2) {Bird};
    \node (D) at (-2,-4) {Dog};
    \node (E) at (2,-4) {Penguin};
    
    \draw[->] (A) -- (B);
    \draw[->] (A) -- (C);
    \draw[->] (B) -- (D);
    \draw[->] (C) -- (E);
\end{tikzpicture}
\end{center}

In the above diagram:

\begin{itemize}
    \item \texttt{Animal} is the base class.
    \item \texttt{Mammal} and \texttt{Bird} are derived classes from \texttt{Animal}.
    \item \texttt{Dog} is derived from \texttt{Mammal}, and \texttt{Penguin} is derived from \texttt{Bird}.
\end{itemize}

This hierarchical structure illustrates the relationships created through inheritance in Python.

\section{Polymorphism}
\subsection{Introduction}
Polymorphism \cite{issariyakul2012review} is one of the core concepts in Object-Oriented Programming (OOP). It refers to the ability of different classes to provide a unique implementation of the same method. The term polymorphism is derived from Greek words meaning "many forms". In Python, polymorphism allows methods to perform different functions based on the object that is calling them. This flexibility promotes code extensibility and reusability, as it allows objects of different classes to respond to the same message (method call) in different ways.

There are two main types of polymorphism:
\begin{itemize}
    \item Compile-time polymorphism (also known as static polymorphism)
    \item Run-time polymorphism (also known as dynamic polymorphism)
\end{itemize}

We will explore these types of polymorphism and how Python implements them, along with detailed examples to help clarify these concepts.

\subsection{Compile-Time Polymorphism (Method Overloading)}
Compile-time polymorphism occurs when multiple methods share the same name but differ in the number or types of parameters. This concept is called method overloading. In languages like Java or C++, you can define multiple methods with the same name but different signatures (parameter types or numbers).

However, Python does not support method overloading in the same way as other languages do. In Python, if you define two methods with the same name, the last method defined will overwrite the previous one. But you can achieve similar functionality by using default parameters or by checking the types and number of arguments inside a single method.

Here’s an example of how you might simulate method overloading in Python:

\begin{lstlisting}[style=python]
class Calculator:
    # Single method handling different numbers of arguments
    def add(self, a, b, c=None):
        if c:
            return a + b + c
        else:
            return a + b

# Creating an instance of Calculator
calc = Calculator()

# Calling 'add' method with two arguments
print(calc.add(10, 20))  # Output: 30

# Calling 'add' method with three arguments
print(calc.add(10, 20, 30))  # Output: 60
\end{lstlisting}

In this example, the method \texttt{add} is capable of handling both two and three arguments. Even though Python does not allow traditional method overloading, we can use default parameters or \texttt{if} statements to achieve similar behavior.

\subsection{Run-Time Polymorphism (Method Overriding)}
Run-time polymorphism is a key feature of OOP, allowing a subclass to provide a specific implementation of a method that is already defined in its parent class. This process is known as method overriding. The implementation of the method that is called is determined at runtime, based on the type of the object.

In Python, you can easily override methods by defining the same method name in a child class, which will override the parent class's method.

Here’s an example demonstrating method overriding:

\begin{lstlisting}[style=python]
class Animal:
    def sound(self):
        return "Some generic sound"

class Dog(Animal):
    def sound(self):
        return "Bark"

class Cat(Animal):
    def sound(self):
        return "Meow"

# Creating instances of Dog and Cat
dog = Dog()
cat = Cat()

# Calling the overridden 'sound' method
print(dog.sound())  # Output: Bark
print(cat.sound())  # Output: Meow
\end{lstlisting}

In this example, both the \texttt{Dog} and \texttt{Cat} classes override the \texttt{sound} method of the \texttt{Animal} class. The method that gets executed depends on the actual object type at runtime, which makes this an example of run-time polymorphism.

\subsection{Dynamic Binding}
Dynamic binding \cite{dewar2006safety} is closely related to run-time polymorphism. It refers to the process by which the method that is to be executed is determined at runtime based on the actual object type. In Python, all method calls are dynamically bound because the method to be invoked is determined by the type of the object at runtime.

Here’s an example to illustrate dynamic binding:

\begin{lstlisting}[style=python]
class Shape:
    def area(self):
        pass

class Circle(Shape):
    def __init__(self, radius):
        self.radius = radius

    def area(self):
        return 3.14 * (self.radius ** 2)

class Rectangle(Shape):
    def __init__(self, width, height):
        self.width = width
        self.height = height

    def area(self):
        return self.width * self.height

# Function that calculates area
def print_area(shape):
    print("Area:", shape.area())

# Creating instances
circle = Circle(5)
rectangle = Rectangle(4, 6)

# Calling print_area with different shapes
print_area(circle)  # Output: Area: 78.5
print_area(rectangle)  # Output: Area: 24
\end{lstlisting}

In this example, \texttt{print\_area} takes a \texttt{Shape} object as input, but the actual method that gets called depends on whether the object is an instance of \texttt{Circle} or \texttt{Rectangle}. This is a clear demonstration of dynamic binding, where the correct \texttt{area} method is determined at runtime.

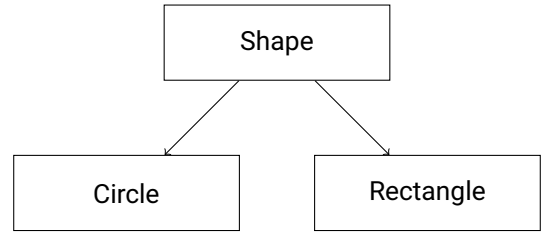
\begin{center}
\begin{tikzpicture}
\tikzstyle{every node}=[draw,rectangle,minimum width=3cm,minimum height=1cm] 
\node (A) at (0,0) {Shape};
\node (B) at (-2,-2) {Circle};
\node (C) at (2,-2) {Rectangle};

\draw [->] (A) -- (B);
\draw [->] (A) -- (C);
\end{tikzpicture}
\end{center}

As shown in the diagram above, both \texttt{Circle} and \texttt{Rectangle} inherit from the \texttt{Shape} class, but each implements the \texttt{area} method differently. Dynamic binding ensures that the correct method is invoked depending on the actual type of the object at runtime.

\subsection{Conclusion}
Polymorphism in Python, both at compile-time and run-time, allows for flexible and reusable code. While Python doesn’t support traditional method overloading like some other languages, you can simulate it using default arguments or argument checking. Run-time polymorphism, achieved through method overriding and dynamic binding, is a powerful feature in Python, allowing for object-specific method implementations to be determined at runtime.

This combination of compile-time flexibility and run-time adaptability makes polymorphism a cornerstone of effective object-oriented programming in Python.

\section{Abstraction}

\subsection{Introduction}
Abstraction \cite{lieberman2006continuing} is a fundamental concept in Object-Oriented Programming (OOP) that focuses on hiding the complex internal implementation details of a class or object and only exposing the necessary and relevant features to the user. This helps reduce complexity and allows the programmer to focus on interacting with the object rather than worrying about the inner workings of its methods or attributes.

In Python, abstraction can be achieved through abstract classes and interfaces. These techniques are essential in software development as they enable code reusability, improve code readability, and make large systems easier to manage by emphasizing the "what" rather than the "how."

\subsection{Abstract Classes}
Abstract classes serve as templates for other classes. They define a blueprint for other classes to inherit but cannot be instantiated directly. In Python, abstract classes are created using the \texttt{abc} module (short for Abstract Base Classes).

An abstract class may have one or more abstract methods, which are methods declared but contain no implementation. Subclasses of the abstract class must provide implementations for these abstract methods. This ensures that any derived class will implement the required functionality while allowing flexibility in how that functionality is provided.

\paragraph{Example: Abstract Class in Python}

Below is an example that demonstrates how abstract classes work in Python. Here, we define an abstract class \texttt{Animal} with an abstract method \texttt{make\_sound}. Any subclass of \texttt{Animal} must implement the \texttt{make\_sound} method.

\begin{lstlisting}[style=python]
from abc import ABC, abstractmethod

class Animal(ABC):
    @abstractmethod
    def make_sound(self):
        pass

class Dog(Animal):
    def make_sound(self):
        return "Bark"

class Cat(Animal):
    def make_sound(self):
        return "Meow"
        
# Creating objects of Dog and Cat
dog = Dog()
cat = Cat()

print(dog.make_sound())  # Output: Bark
print(cat.make_sound())  # Output: Meow
\end{lstlisting}

In this example, \texttt{Animal} is the abstract class with the abstract method \texttt{make\_sound}. Both \texttt{Dog} and \texttt{Cat} are concrete subclasses that implement the \texttt{make\_sound} method.

Attempting to instantiate the \texttt{Animal} class directly will result in an error, as it contains an abstract method that has not been implemented. This ensures that each subclass provides its specific implementation.

\subsection{Interfaces}
In Python, interfaces can be created using abstract base classes (ABCs). An interface is a type of abstract class that defines a set of methods that a class must implement, but it does not provide any implementation itself.

By using interfaces, you can enforce method contracts. This means that any class implementing the interface must provide concrete implementations for all the abstract methods defined in the interface. In Python, an interface is effectively just an abstract class where every method is abstract.

\paragraph{Example: Interface in Python}

Consider the following example, which illustrates how to define and use an interface in Python:

\begin{lstlisting}[style=python]
from abc import ABC, abstractmethod

class Shape(ABC):
    @abstractmethod
    def area(self):
        pass

    @abstractmethod
    def perimeter(self):
        pass

class Rectangle(Shape):
    def __init__(self, width, height):
        self.width = width
        self.height = height

    def area(self):
        return self.width * self.height

    def perimeter(self):
        return 2 * (self.width + self.height)

class Circle(Shape):
    def __init__(self, radius):
        self.radius = radius

    def area(self):
        return 3.14159 * self.radius ** 2

    def perimeter(self):
        return 2 * 3.14159 * self.radius
        
# Creating objects of Rectangle and Circle
rect = Rectangle(10, 20)
circ = Circle(5)

print(rect.area())       # Output: 200
print(rect.perimeter())  # Output: 60
print(circ.area())       # Output: 78.53975
print(circ.perimeter())  # Output: 31.4159
\end{lstlisting}

In this example, \texttt{Shape} is an abstract class acting as an interface that defines two abstract methods: \texttt{area} and \texttt{perimeter}. Both \texttt{Rectangle} and \texttt{Circle} classes implement these methods according to their specific geometrical formulas.

\subsection{Use Cases of Abstraction}
Abstraction is widely used in real-world programming to simplify complex systems and improve code maintainability. Below are some practical use cases where abstraction plays a crucial role:

\paragraph{1. GUI Applications:}
In graphical user interface (GUI) programming, abstraction is often used to manage different user interface components. For example, frameworks like Tkinter or PyQt provide abstract classes for buttons, text fields, and windows. Developers can use these classes without worrying about how the elements are rendered on different operating systems.

\begin{lstlisting}[style=python]
import tkinter as tk

class MyApp:
    def __init__(self, root):
        self.button = tk.Button(root, text="Click Me", command=self.say_hello)
        self.button.pack()

    def say_hello(self):
        print("Hello, World!")

root = tk.Tk()
app = MyApp(root)
root.mainloop()
\end{lstlisting}

In the example above, we do not need to know how the button is drawn on the screen. The framework abstracts this complexity for us, and we only interact with the essential features, such as defining the button's text and functionality.

\paragraph{2. Database Connectivity:}
Abstraction is also applied in database management systems (DBMS). Using object-relational mappers (ORMs) like SQLAlchemy or Django ORM, developers work with Python objects instead of directly writing SQL queries. The ORM abstracts the details of the database and provides a clean API for interacting with it.

\begin{lstlisting}[style=python]
from sqlalchemy import create_engine, Column, Integer, String, Base

engine = create_engine('sqlite:///:memory:')

class User(Base):
    __tablename__ = 'users'
    id = Column(Integer, primary_key=True)
    name = Column(String)

Base.metadata.create_all(engine)
\end{lstlisting}

Here, the \texttt{User} class represents a table in the database. SQLAlchemy abstracts away the need to write complex SQL queries, allowing the developer to focus on the higher-level application logic.

\paragraph{3. File Handling:}
Python's built-in file handling methods abstract the complexity of reading and writing data from files. For example, you can open, read, write, and close a file without needing to manage low-level details like file pointers or system calls.

\begin{lstlisting}[style=python]
# Opening a file for reading
with open('example.txt', 'r') as file:
    content = file.read()

print(content)

# File is automatically closed after the block
\end{lstlisting}

The \texttt{with} statement abstracts the file opening and closing process. This not only simplifies the code but also ensures that resources are handled efficiently.

\subsection{Conclusion}
Abstraction is a powerful concept that helps simplify complex systems and makes the code more manageable and maintainable. By using abstract classes and interfaces, developers can create flexible, reusable, and well-structured code that is easier to understand and modify. Whether in GUI applications, database management, or file handling, abstraction plays a critical role in reducing complexity and improving overall software quality.

\section{Association, Aggregation, and Composition}
    \subsection{Introduction}
    This section explores three important types of relationships between objects in object-oriented programming (OOP): association, aggregation, and composition. These relationships define how objects interact and relate to each other, which is crucial in designing well-structured programs. Each type of relationship has its own characteristics, and understanding these differences will help in modeling real-world relationships effectively in Python.

    \subsection{Association}
    Association \cite{williams2002associative} is a general term used to define a relationship between two objects. In association, one object uses or interacts with another object without owning it. It’s a "has-a" relationship, but the lifetime of the associated objects is independent. This is the most basic form of relationship between objects. For example, a teacher might be associated with multiple students, and a student might be associated with multiple teachers, but neither owns the other.

    In Python, association is typically represented by making one object a parameter or an attribute in another object. Here is an example of association in Python:

    \begin{lstlisting}[style=python]
class Teacher:
    def __init__(self, name):
        self.name = name

    def teach(self):
        return f"{self.name} is teaching."

class Student:
    def __init__(self, name):
        self.name = name

    def learn(self):
        return f"{self.name} is learning."

# Creating objects
teacher1 = Teacher("Mrs. Smith")
student1 = Student("John")

# Association: Teacher interacts with Student, but they don't own each other
print(teacher1.teach())
print(student1.learn())
    \end{lstlisting}

    In this example, the `Teacher` and `Student` classes have an association, as the teacher teaches and the student learns, but there’s no ownership between them.

    \subsection{Aggregation}
    Aggregation \cite{johnson1993refactoring} is a specialized form of association where one object contains another, but the contained object can exist independently of the container. This is still a "has-a" relationship, but in this case, the lifetime of the contained object is not dependent on the container object. In other words, if the container object is destroyed, the contained object can still exist.

    For example, a school can have teachers, but the teachers can still exist without the school. Aggregation allows for this kind of flexibility in object relationships. Below is an example of aggregation in Python:

    \begin{lstlisting}[style=python]
class School:
    def __init__(self, name):
        self.name = name
        self.teachers = []

    def add_teacher(self, teacher):
        self.teachers.append(teacher)

    def get_teachers(self):
        return [teacher.name for teacher in self.teachers]

class Teacher:
    def __init__(self, name):
        self.name = name

# Teachers can exist independently of the school
teacher1 = Teacher("Mrs. Smith")
teacher2 = Teacher("Mr. Johnson")

# School aggregates teachers
school = School("Greenwood High")
school.add_teacher(teacher1)
school.add_teacher(teacher2)

print(f"Teachers at {school.name}: {school.get_teachers()}")
    \end{lstlisting}

    In this example, `School` contains multiple `Teacher` objects, but the `Teacher` objects can exist outside the school. This demonstrates aggregation, as the `Teacher` objects can exist independently of the `School` object.

    \subsection{Composition}
    Composition \cite{nierstrasz1995object} is a stronger form of aggregation where one object owns another object. In this case, the contained object cannot exist without the container. This implies a strict lifecycle dependency between the container and the contained object. When the container object is destroyed, the contained object is also destroyed.

    For instance, a university can have departments, and when the university is dissolved, the departments cease to exist as well. Below is an example of composition in Python:

    \begin{lstlisting}[style=python]
class University:
    def __init__(self, name):
        self.name = name
        self.departments = []

    def add_department(self, department_name):
        department = self.Department(department_name)
        self.departments.append(department)

    class Department:
        def __init__(self, name):
            self.name = name

    def get_departments(self):
        return [department.name for department in self.departments]

# University and departments have a composition relationship
university = University("Harvard University")
university.add_department("Computer Science")
university.add_department("Mathematics")

print(f"Departments at {university.name}: {university.get_departments()}")
    \end{lstlisting}

    In this example, the `Department` class is nested within the `University` class. This reflects a composition relationship because departments cannot exist independently of the university. If the `University` object is deleted, the `Department` objects would also be deleted automatically.

    \subsection{Differences between Aggregation and Composition}
    While both aggregation and composition represent "has-a" relationships, the key difference lies in the lifecycle dependency between the contained and container objects:
    
    \begin{itemize}
        \item \textbf{Aggregation:} The contained object can exist independently of the container object. For example, a `Teacher` can exist without a `School`.
        \item \textbf{Composition:} The contained object cannot exist independently of the container object. For example, a `Department` cannot exist without a `University`.
    \end{itemize}
    
    The following diagram illustrates the differences between association, aggregation, and composition in object relationships:

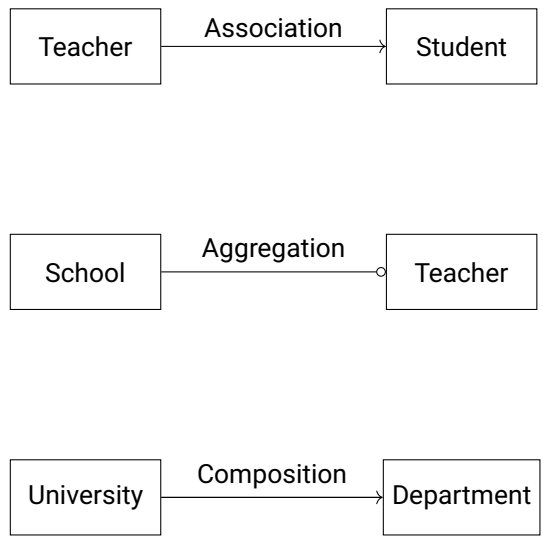
\begin{center}
\begin{tikzpicture}[node distance=2cm]
    \usetikzlibrary{arrows.meta}

    \node (assoc1) [rectangle, draw, minimum width=2cm, minimum height=1cm] {Teacher};
    \node (assoc2) [rectangle, draw, right of=assoc1, xshift=3cm, minimum width=2cm, minimum height=1cm] {Student};
    \draw[->] (assoc1) -- (assoc2) node[midway, above] {Association};
    
    \node (agg1) [rectangle, draw, below of=assoc1, yshift=-1cm, minimum width=2cm, minimum height=1cm] {School};
    \node (agg2) [rectangle, draw, right of=agg1, xshift=3cm, minimum width=2cm, minimum height=1cm] {Teacher};
    \draw[-{Circle[open]}] (agg1) -- (agg2) node[midway, above] {Aggregation};
    
    \node (comp1) [rectangle, draw, below of=agg1, yshift=-1cm, minimum width=2cm, minimum height=1cm] {University};
    \node (comp2) [rectangle, draw, right of=comp1, xshift=3cm, minimum width=2cm, minimum height=1cm] {Department};
    \draw[->,fill] (comp1) -- (comp2) node[midway, above] {Composition};
\end{tikzpicture}
\end{center}

    This diagram shows that in association, there is a simple interaction between objects. In aggregation, the container object can exist without the contained object, while in composition, the contained object cannot exist independently of the container.

    Understanding these relationships is crucial for designing complex systems in Python, as they help you model real-world relationships between different entities effectively and organize your code in a maintainable way.

\section{Design Principles in OOP}
    \subsection{Introduction}
    When working with object-oriented programming (OOP), particularly in Python, it is essential to follow design principles that help produce clean, efficient, and maintainable code. These principles guide developers in structuring code that is easy to understand, modify, and extend without introducing errors or unnecessary complexity. Some of the most widely accepted design principles are the SOLID principles, along with others like DRY and KISS \cite{russell2024principles}. Each of these principles ensures that your code remains organized and reduces the risk of introducing bugs during development or future changes.

    \subsection{SOLID Principles}
    The SOLID principles \cite{madasu2015solid} are a set of five design principles that, when followed, enable developers to write better, more maintainable, and extensible object-oriented software. Let’s dive into each principle and explore it in the context of Python with clear examples.

        \subsubsection{Single Responsibility Principle}
        The Single Responsibility Principle (SRP) \cite{holderer2022single} states that a class should have only one reason to change, meaning that it should have only one responsibility. In simple terms, a class should do one thing and do it well. If a class has more than one responsibility, it becomes harder to maintain and modify without affecting other functionalities.

        \textbf{Example:} Let's consider a class that manages both user data and logging functionality. This would violate SRP.

\begin{lstlisting}[style=python]
class UserManager:
    def __init__(self, user_data):
        self.user_data = user_data
    
    def save_user(self):
        # Code to save user data
        pass
    
    def log_action(self, message):
        # Code to log an action
        print(f"Log: {message}")
\end{lstlisting}

        In this case, the class handles both user management and logging. To adhere to SRP, we should split these responsibilities into two classes:

\begin{lstlisting}[style=python]
class UserManager:
    def __init__(self, user_data):
        self.user_data = user_data
    
    def save_user(self):
        # Code to save user data
        pass

class Logger:
    def log_action(self, message):
        print(f"Log: {message}")
\end{lstlisting}

        Now, each class has only one responsibility: \texttt{UserManager} manages user data, and \texttt{Logger} handles logging.

        \subsubsection{Open/Closed Principle}
        The Open/Closed Principle (OCP) \cite{skoglund2003practical} dictates that software entities (such as classes, modules, or functions) should be open for extension but closed for modification. This means that you should be able to extend a class's behavior without modifying its existing code, which reduces the risk of introducing bugs.

        \textbf{Example:} Suppose we have a class that calculates the area of different shapes:

\begin{lstlisting}[style=python]
class AreaCalculator:
    def calculate_area(self, shape):
        if shape == "circle":
            return 3.14 * 5 * 5
        elif shape == "rectangle":
            return 10 * 20
\end{lstlisting}

        This class violates OCP because any time we need to add support for a new shape, we must modify the \texttt{calculate\_area} method. To adhere to OCP, we can use inheritance and polymorphism:

\begin{lstlisting}[style=python]
class Shape:
    def calculate_area(self):
        pass

class Circle(Shape):
    def __init__(self, radius):
        self.radius = radius
    
    def calculate_area(self):
        return 3.14 * self.radius * self.radius

class Rectangle(Shape):
    def __init__(self, width, height):
        self.width = width
        self.height = height
    
    def calculate_area(self):
        return self.width * self.height

class AreaCalculator:
    def calculate_area(self, shape: Shape):
        return shape.calculate_area()
\end{lstlisting}

        Now, the code is open for extension (new shapes can be added easily by creating new classes) but closed for modification (we don’t need to change the \texttt{AreaCalculator} class).

        \subsubsection{Liskov Substitution Principle}
        The Liskov Substitution Principle (LSP) \cite{haoyu2012basic} states that objects of a subclass should be able to replace objects of the superclass without affecting the correctness of the program. This principle encourages us to ensure that subclasses behave in a way that their use in place of a parent class doesn't break the functionality.

        \textbf{Example:} Let's consider a simple hierarchy of animals:

\begin{lstlisting}[style=python]
class Bird:
    def fly(self):
        print("Bird is flying")

class Penguin(Bird):
    def fly(self):
        raise Exception("Penguins can't fly")
\end{lstlisting}

        In this case, if you replace \texttt{Bird} with \texttt{Penguin}, it breaks the program since a penguin cannot fly. This violates LSP. To follow LSP, we need to ensure that substituting a subclass does not introduce such contradictions:

\begin{lstlisting}[style=python]
class Bird:
    def move(self):
        pass

class FlyingBird(Bird):
    def move(self):
        print("Flying")

class Penguin(Bird):
    def move(self):
        print("Swimming")
\end{lstlisting}

        Now, both \texttt{FlyingBird} and \texttt{Penguin} can be substituted for \texttt{Bird} without breaking the program's correctness.

        \subsubsection{Interface Segregation Principle}
        The Interface Segregation Principle (ISP) \cite{madasu2015solid} suggests that no client should be forced to depend on methods it does not use. In Python, where explicit interfaces are not used like in other languages, this principle can be applied by ensuring that classes don't have unnecessary methods that are irrelevant to certain clients.

        \textbf{Example:} Suppose we have a \texttt{Worker} class:

\begin{lstlisting}[style=python]
class Worker:
    def work(self):
        pass
    
    def eat(self):
        pass
\end{lstlisting}

        Now, if we create a \texttt{RobotWorker} class that inherits from \texttt{Worker}, it would inherit the \texttt{eat} method, which makes no sense for a robot. This violates ISP. To fix this, we can separate the interfaces:

\begin{lstlisting}[style=python]
class Workable:
    def work(self):
        pass

class Eatable:
    def eat(self):
        pass

class HumanWorker(Workable, Eatable):
    def work(self):
        print("Working")
    
    def eat(self):
        print("Eating")

class RobotWorker(Workable):
    def work(self):
        print("Working")
\end{lstlisting}

        Now, each class only depends on the methods it needs.

        \subsubsection{Dependency Inversion Principle}
        The Dependency Inversion Principle (DIP) \cite{thennakoon2022study} states that high-level modules should not depend on low-level modules but both should depend on abstractions. This means that the details of implementation should depend on abstractions (interfaces) rather than the other way around.

        \textbf{Example:} Consider a class that directly depends on a low-level class:

\begin{lstlisting}[style=python]
class LightBulb:
    def turn_on(self):
        print("LightBulb turned on")

class Switch:
    def __init__(self, bulb: LightBulb):
        self.bulb = bulb
    
    def operate(self):
        self.bulb.turn_on()
\end{lstlisting}

        Here, \texttt{Switch} depends directly on \texttt{LightBulb}, violating DIP. To follow DIP, we introduce an abstraction:

\begin{lstlisting}[style=python]
class Switchable:
    def turn_on(self):
        pass

class LightBulb(Switchable):
    def turn_on(self):
        print("LightBulb turned on")

class Fan(Switchable):
    def turn_on(self):
        print("Fan turned on")

class Switch:
    def __init__(self, device: Switchable):
        self.device = device
    
    def operate(self):
        self.device.turn_on()
\end{lstlisting}

        Now, \texttt{Switch} depends on the abstraction \texttt{Switchable}, allowing us to easily swap out different devices like \texttt{Fan} or \texttt{LightBulb} without modifying the \texttt{Switch} class.

    \subsection{DRY (Don't Repeat Yourself)}
    The DRY principle \cite{keey2013drying} emphasizes avoiding repetition of code or logic. If you find yourself writing the same or similar code in multiple places, it’s better to abstract it into a function, class, or module that can be reused. This improves code maintainability and reduces the likelihood of errors.

    \textbf{Example:} Let’s say we calculate the area of different shapes in multiple parts of the code:

\begin{lstlisting}[style=python]
# Code block 1
area_of_circle = 3.14 * radius * radius

# Code block 2
area_of_rectangle = width * height
\end{lstlisting}

    To adhere to the DRY principle, we can refactor this into a function:

\begin{lstlisting}[style=python]
def calculate_area(shape, **dimensions):
    if shape == "circle":
        return 3.14 * dimensions['radius'] * dimensions['radius']
    elif shape == "rectangle":
        return dimensions['width'] * dimensions['height']
\end{lstlisting}

    Now the code is more maintainable and reusable.

    \subsection{KISS (Keep It Simple, Stupid)}
    The KISS principle \cite{alwin2016kiss} states that systems should be as simple as possible. Avoid overcomplicating solutions and aim to write clear, straightforward code that is easy to understand. Unnecessary complexity can lead to confusion, increased bugs, and difficulties when maintaining or extending the system.

    \textbf{Example:} Instead of over-engineering a solution for a simple task like finding the maximum of two numbers:

\begin{lstlisting}[style=python]
def find_max(a, b):
    if a > b:
        return a
    else:
        return b
\end{lstlisting}

    A simpler and more readable approach is:

\begin{lstlisting}[style=python]
def find_max(a, b):
    return a if a > b else b
\end{lstlisting}

    Keeping things simple makes the code easier to maintain and reduces the chance of errors.

\section{UML (Unified Modeling Language)}

\subsection{Introduction}
Unified Modeling Language (UML) \cite{booch1996unified} is a standardized modeling language used to visualize the design and structure of software systems. It provides a variety of diagram types that help in documenting, planning, and understanding the architecture and behavior of a system. UML is particularly useful for Object-Oriented Programming (OOP) because it allows us to model classes, objects, and their interactions \cite{alhir1999understanding}. In this section, we will explore key UML diagrams used in software development, focusing on how they relate to Python OOP concepts. These diagrams include class diagrams, object diagrams, sequence diagrams, and use case diagrams.

\subsection{Class Diagrams}
Class diagrams \cite{berardi2005reasoning} are fundamental to understanding Object-Oriented Programming. They show the structure of a system by illustrating its classes, their attributes (data), methods (functions), and the relationships between them. Class diagrams help developers visualize how different parts of a system are connected and interact with one another.

In Python, a class diagram maps directly to the `class` keyword, which is used to define a blueprint for creating objects. Let’s consider a simple Python class for a `Car`, and show how this would be represented in a UML class diagram.

\begin{lstlisting}[style=python]
class Car:
    def __init__(self, make, model, year):
        self.make = make
        self.model = model
        self.year = year

    def start(self):
        print(f"{self.make} {self.model} is starting.")

    def stop(self):
        print(f"{self.make} {self.model} is stopping.")
\end{lstlisting}

In UML, this class would be represented as a box with three compartments. The top compartment contains the class name (`Car'), the middle contains the attributes (`make', `model', `year'), and the bottom contains the methods (`start()', `stop()').

Here is a basic UML class diagram for the `Car' class: \\

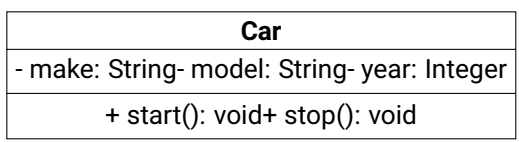
\usetikzlibrary{shapes.multipart} 
\begin{tikzpicture}
  \node[draw, rectangle split, rectangle split parts=3] at (0,0) {
    \textbf{Car}
    \nodepart{second}
    - make: String \\
    - model: String \\
    - year: Integer
    \nodepart{third}
    + start(): void \\
    + stop(): void
  };
\end{tikzpicture} \\

The `-` symbol indicates private attributes, while the `+` symbol represents public methods.

\subsection{Object Diagrams}
While class diagrams show the static structure of classes, object diagrams \cite{maoz2011modal} provide a snapshot of instances (objects) of those classes at a specific moment in time. An object diagram demonstrates how objects interact and what data they contain at runtime.

Consider an instance of the `Car` class in Python:

\begin{lstlisting}[style=python]
my_car = Car("Toyota", "Corolla", 2020)
\end{lstlisting}

In UML, an object diagram would show an instance of the `Car` class, such as `my\_car`, with its current state. \\

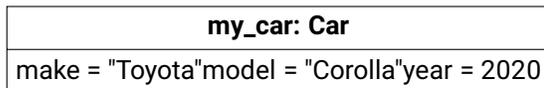
\begin{tikzpicture}
  \node[draw, rectangle split, rectangle split parts=2] at (0,0) {
    \textbf{my\_car: Car}
    \nodepart{second}
    make = "Toyota" \\
    model = "Corolla" \\
    year = 2020
  };
\end{tikzpicture} \\

This shows the current state of the `my\_car` object at a particular point in the program's execution.

\subsection{Sequence Diagrams}
Sequence diagrams \cite{maoz2011modal} focus on the interactions between objects over time. They show how objects communicate with each other, particularly the order of method calls. Sequence diagrams are helpful for understanding how different parts of the system work together to achieve specific tasks.

Let’s consider the following Python interaction between a `Driver` and a `Car` object:

\begin{lstlisting}[style=python]
class Driver:
    def __init__(self, name, car):
        self.name = name
        self.car = car

    def start_trip(self):
        self.car.start()

    def stop_trip(self):
        self.car.stop()

car = Car("Toyota", "Corolla", 2020)
driver = Driver("Alice", car)

driver.start_trip()
driver.stop_trip()
\end{lstlisting}

In this example, `Driver` interacts with `Car` by calling its methods `start()` and `stop()`. The sequence diagram will show this interaction over time, beginning with the `start\_trip()` method and ending with the `stop\_trip()` method. \\

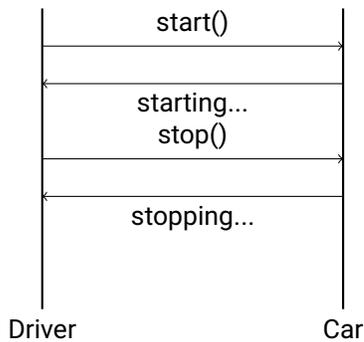
\begin{tikzpicture}
\draw[thick] (0,0) -- (0,-4) node[below] {Driver};
\draw[thick] (4,0) -- (4,-4) node[below] {Car};

\draw[->] (0,-0.5) -- (4,-0.5) node[midway, above] {start()};
\draw[->] (4,-1) -- (0,-1) node[midway, below] {starting...};

\draw[->] (0,-2) -- (4,-2) node[midway, above] {stop()};
\draw[->] (4,-2.5) -- (0,-2.5) node[midway, below] {stopping...};
\end{tikzpicture} \\

This sequence diagram clearly shows the communication between the `Driver` and `Car` objects, including the order of method calls.

\subsection{Use Case Diagrams}
Use case diagrams \cite{wegmann2000role} represent the functional requirements of a system. They focus on how different users (called actors) interact with the system. Use case diagrams help in understanding the behavior of a system from a user’s perspective.

In a simple system like a car rental service, we might have two actors: a `Customer` and a `System`. The customer interacts with the system by performing actions such as searching for available cars, booking a car, and returning a car. These actions are represented as use cases.

Here’s a basic use case diagram: \\

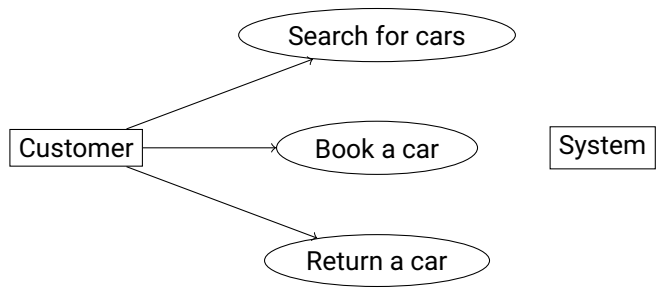
\usetikzlibrary{shapes.geometric} 
\begin{tikzpicture}
  \node[draw] (system) at (3,0) {System};
  \node[draw, ellipse] (search) at (0, 1.5) {Search for cars};
  \node[draw, ellipse] (book) at (0, 0) {Book a car};
  \node[draw, ellipse] (return) at (0, -1.5) {Return a car};
  \node[draw] (customer) at (-4, 0) {Customer};

  \draw[->] (customer) -- (search);
  \draw[->] (customer) -- (book);
  \draw[->] (customer) -- (return);
\end{tikzpicture} \\

In this use case diagram, the `Customer` actor interacts with the `System` by performing various use cases (e.g., searching for cars, booking a car, and returning a car). This type of diagram helps clarify what functionality the system needs to provide for its users.

\bibliographystyle{ieeetr}
\bibliography{sample}

\end{document}